# A Robust System for Natural Spoken Dialogue

James F. Allen, Bradford W. Miller, Eric K. Ringger, Teresa Sikorski

Dept. of Computer Science
University of Rochester
Rochester, NY 14627
{james, miller, ringger, sikorski}@cs.rochester.edu
http://www.cs.rochester.edu/research/trains/

## Abstract

This paper describes a system that leads us to believe in the feasibility of constructing natural spoken dialogue systems in task-oriented domains. It specifically addresses the issue of robust interpretation of speech in the presence of recognition errors. Robustness is achieved by a combination of statistical error post-correction, syntactically- and semantically-driven robust parsing, and extensive use of the dialogue context. We present an evaluation of the system using time-to-completion and the quality of the final solution that suggests that most native speakers of English can use the system successfully with virtually no training.

## 1. Introduction

While there has been much research on natural dialogue, there have been few working systems because of the difficulties in obtaining robust behavior. Given over twenty years of research in this area, if we can't construct a robust system even in a simple domain then that bodes ill for progress in the field. In particular, without some working systems, we are very limited in how we can evaluate the worth of different models.

The prime goal of the work reported here was to demonstrate that it is feasible to construct robust spoken natural dialogue systems. We were not seeking to develop new theories, but rather to develop techniques to enable existing theories to be applied in practice. We chose a domain and task that was as simple as possible yet couldn't be solved without the collaboration of the human and system. In addition, there were three fundamental requirements:
- the system must run in near real-time;
- the user should need minimal training and not be constrained in what can be said; and
- the dialogue should have a concrete result that can be independently evaluated.

The second constraint means we must handle natural dialogue, namely dialogue as people use it rather than a constrained form of interaction determined by the system (which is often called a dialogue). We can only control the complexity of the dialogue by controlling the complexity of the task. Increasing the task complexity naturally increases the complexity of the dialogue. This paper reports on the first stage of this process, working with a highly simplified domain.

At the start of this experiment in November 1994, we had no idea whether it was possible. While researchers were reporting good accuracy (upwards of 95%) for speech systems in simple question-answering tasks, our domain was considerably different with a much more spontaneous form of interaction.

We also knew that it would not be possible to directly use general models of plan recognition to aid in speech act interpretation (as in Allen & Perrault, 1980, Litman & Allen 1987, Carberry 1990), as these models would not lend themselves to real-time processing. Similarly, it would not be feasible to use general planning models for the system back-end and for planning its responses. We could not, on the other hand, completely abandon the ideas underlying the plan-based approach, as we knew of no other theory that could provide an account for the interactions. Our approach was to try to retain the overall structure of plan-based systems, but to use domain-specific reasoning techniques to provide real-time performance.

Dialogue systems are notoriously hard to evaluate as there is no well-defined "correct answer". So we cannot give end-to-end accuracy measures as is typically done to measure the performance of speech recognition systems and parsing systems. This is especially true when evaluating dialogue robustness, which results from many different sources: correcting speech recognition errors, using semantic knowledge to interpret fragments, and using dialogue strategies to keep the dialogue flowing efficiently despite recognition and interpretation errors.

The approach we take is to use task-based evaluation. We measure how well the system does at helping the user solve the problem. The two most telling measures are time-to-completion and the quality of the final solution. In the evaluation described later in this paper, we show that all our subjects were able to use TRAINS-95 to solve problems with only minimal training. We also evaluated the overall effectiveness of our robust processing techniques by comparing spoken dialogues with keyboard dialogues by the same subjects. Even with a 30% word error rate (including insertion errors), speech turned out to be considerably more efficient than keyboard input in this domain.

## 2. The Task and Example Session

The domain in TRAINS-95 is simple route planning. The user is given a map on a screen showing cities,

connections and the locations of a set of trains (see Figure 1), and a verbal specification of a set of destination cities where trains are needed. The task is to find the most efficient set of routes possible. The route planner used by the system is deliberately weak so that interaction is needed to find good plans. Specifically, the planner cannot find routes longer than four hops without an intermediate city, and when it can generate a route, it randomly selects among the possibilities.

Typically, the examples in papers on dialogue systems are carefully hand-constructed and designed to show off specific system capabilities. Such examples, however, give little feeling for how robust a system is. Much more revealing is how a system responds to unplanned interactions with naive users, and how it deals with the interaction when things start to go wrong. Thus, we won't present any examples in which everything went smoothly. Rather, we present a dialogue that stresses the system's capabilities, better illuminating what works and what areas require further improvement.

The dialogue below was with a non-computer person who had used the system a few times before. In the transcripts, we show the input as it was produced by the speech recognizer in ALL CAPS followed by what was actually said. If the recognition was perfect we just show the speech recognition output. The dialogue contains a significant number of errors and has a 29.5% word error rate. The system display is also shown at key points[1]. In this problem, the engines start at Detroit, Montreal and Albany and the goal is to get trains to Milwaukee, Lexington and Washington.

S: Ok. I think I'm ready to start.
U: OKAY LET'S SEE CONTAIN FROM DETROIT TO WASHINGTON
   Okay. Let's take a train from Detroit to Washington.
S: What route would you like to get from Detroit to Washington?
U: LET'S GO VIA TOLEDO AND PITTSBURGH
S: *(displays Figure 2)* The terminal at city Scranton is delaying traffic due to localized heavy winds. The terminal at city Baltimore is delaying traffic due to localized heavy winds. An additional five hours will be needed to travel through them due to decreased visibility.
U: NOW LET'S TAKE THE TRAIN FROM DETROIT TO WASHINGTON D S_X ANY
   No. Let's take the train from Detroit to Washington via Cincinnati.
S: *(displays Figure 3)* Okay.
U: OKAY IT THAT'S OKAY NOW
   Okay that - that's okay now
S: Pardon Me?
U: UP OKAY NOW LET'S TAKE THE TRAIN FROM MONTREAL TO LEXINGTON
   Okay. Now let's take the train from Montreal to Lexington.

---
[1] We apologize for the figures. The actual system uses color displays. The best way to present examples is by video. A quicktime movie can be found in

http://www.cs.rochester.edu/research/trains/ACL96.

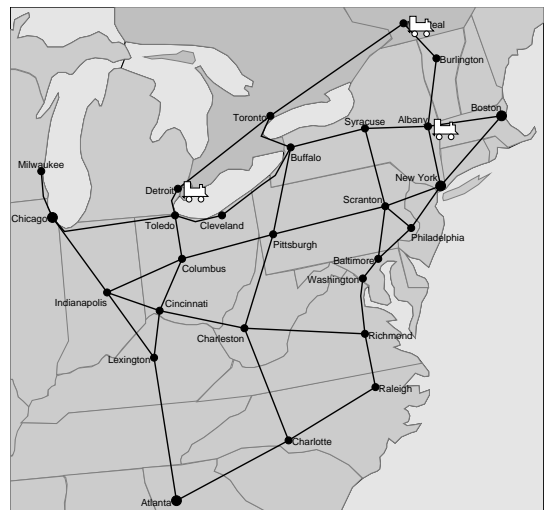
*Figure 1: The initial scenario*

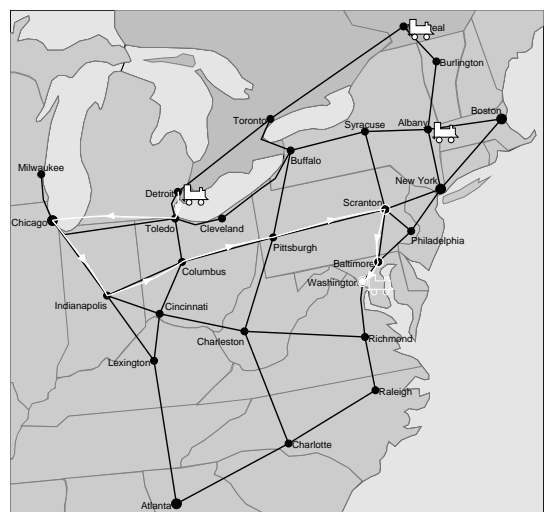
*Figure 2: The proposed route*

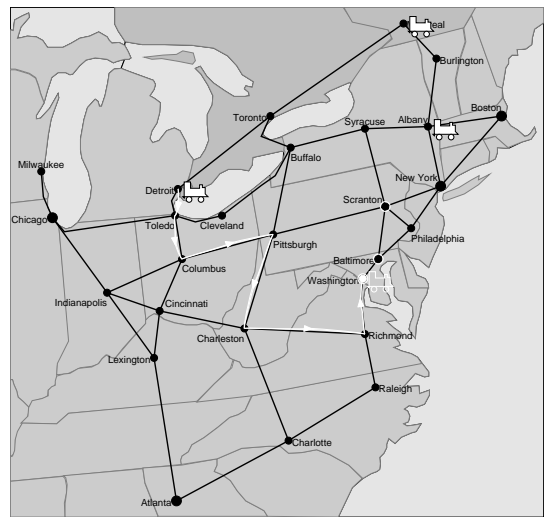
*Figure 3: The corrected route*

*2*

S: I need help choosing a route from Montreal to Lexington.
U: LET'S GO P_M TO TRY
   Let's go via Detroit
S: *(displays a route to Detroit but not on to Lexington)* Is this OK?
U: YES NOW LET'S GO TO LEXINGTON
S: Yep.
U: OKAY NOW ADD STATE THE LAST TRAIN AND GO FROM ALBANY TO O_O'S
   Okay Now let's take the last train and go from Albany to Milwaukee.
S: No. What city are you trying to have the engine at Montreal arrive at, if I caught your drift. I don't understand where we are sending the engine at Albany to.
U: ME A JET ADD ALBANY NEEDS TO GO TO MILWAUKEE
   The engine at Albany needs to go to Milwaukee.
S: Tell me a route to use to get from Albany to Milwaukee, please.
U: I'D GO VIA BUFFALO
   Uh - Go via Buffalo.
S: *(displays a route through Toronto and Buffalo)* Yes.
U: GO B_X SYRACUSE AT BUFFALO
   Go via Syracuse and Buffalo.
S: *(displays Figure 4)* yeah.
U: THAT'S COULD I CAN
   That's good. I'm done.
S: Hey, its the programming.
U: I_NEED DONE
   I'm done.

## 3. The System

The TRAINS-95 system is organized as shown in Figure 5. At the top are the I/O facilities. The speech recognition system is the Sphinx-II system from CMU (Huang et al, 1993). The speech synthesizer is a commercial product: the TRUETALK system from Entropics. The rest of the system was built at Rochester. The display supports a communication language that allows other modules to control the content of the display. It also handles keyboard input. The speech recognition output is passed through the post-processor described in section 4. The parser, described in section 5, accepts input either from the post-processor (for speech) or the display manager (for keyboard), and produces a set of speech act interpretations that are passed to the discourse manager, described in section 6. The discourse manager breaks into a range of subcomponents handling reference, speech act interpretation and planning (the verbal reasoner), and the back-end of the system: the problem solver and domain reasoner. When a speech act is planned for output, it is passed to the generator, which constructs a sentence and passes this to both the speech synthesizer and the display.

The generator is a simple template-based system. It uses templates associated with different speech act forms that are instantiated with descriptions of the particular objects involved. The form of these descriptions is defined for each class of objects in the domain.

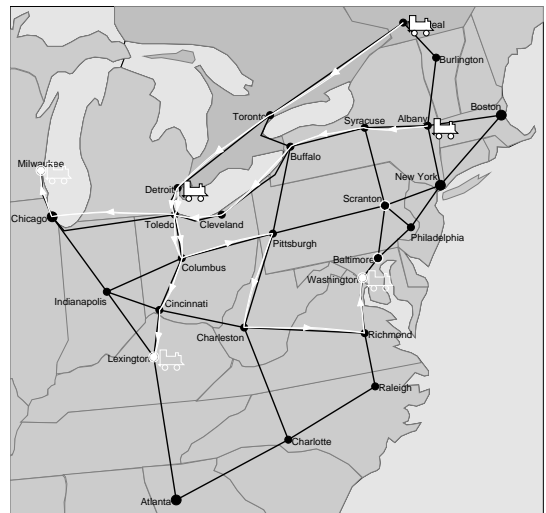
*Figure 4: The final routes*

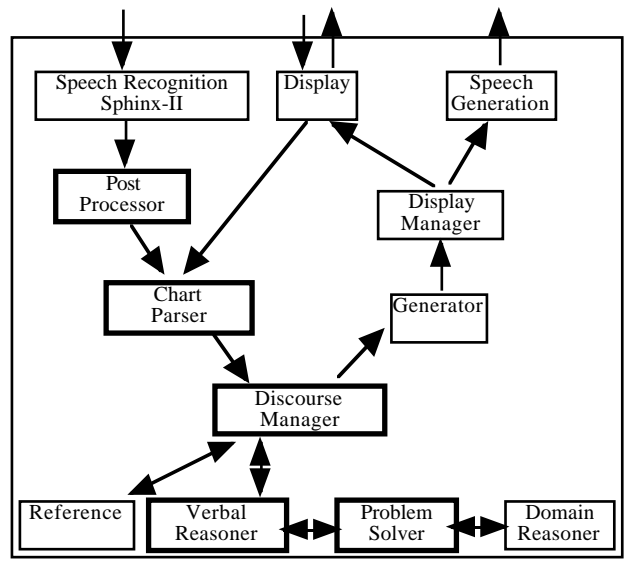
*Figure 5 : The TRAINS-95 System Architecture*

In order to stress the system in our robustness evaluation, we used the ATIS language model provided from CMU. This system yields an overall word error rate of 30% on TRAINS-95 dialogues, as opposed to a 20% error rate that we can currently obtain by using language models trained on our TRAINS corpus. While this accuracy rate is significantly lower than often reported in the literature, remember that most speech recognition results are reported for read speech, or for constrained dialogue applications such as ATIS. Natural dialogue involves a more spontaneous form of interaction that is much more difficult to interpret.

## 4. Statistical Error Post-Correction

The following are examples of speech recognition (SR) errors that occurred in the sample dialogue. In each, the words tagged REF indicate what was actually said, while those tagged with HYP indicate what the speech recognition system proposed, and HYP' indicates the output of SPEECHPP, our post-processor. While the corrected transcriptions are not perfect, they are typically a better approximation of the actual utterance. As the first example shows, some recognition errors are simple word-for-word confusions:

HYP: GO B_X SYRACUSE AT BUFFALO
HYP': GO VIA SYRACUSE VIA BUFFALO
REF: GO VIA SYRACUSE AND BUFFALO

In the next example, a single word was replaced by more than one smaller word:

HYP: LET'S GO P_M TO TRY
HYP': LET'S GO P_M TO DETROIT
REF: LET'S GO VIA DETROIT

The post-processor yields fewer errors by effectively refining and tuning the vocabulary used by the speech recognizer. To achieve this, we adapted some techniques from statistical machine translation (such as Brown et al., 1990) in order to model the errors that Sphinx-II makes in our domain. Briefly, the model consists of two parts: a channel model, which accounts for errors made by the SR, and the language model, which accounts for the likelihood of a sequence of words being uttered in the first place.

More precisely, given an observed word sequence $o$ from the speech recognizer, SPEECHPP finds the most likely original word sequence by finding the sequence $s$ that maximizes $Prob(o|s) * Prob(s)$, where

- $Prob(s)$ is the probability that the user would utter sequence $s$, and
- $Prob(o|s)$ is the probability that the SR produces the sequence $o$ when $s$ was actually spoken.

For efficiency, it is necessary to estimate these distributions with relatively simple models by making independence assumptions. For $Prob(s)$, we train a word-bigram "back-off" language model (Katz, 87) from hand-transcribed dialogues previously collected with the TRAINS-95 system. For $P(o|s)$, we build a channel model that assumes independent word-for-word substitutions; i.e.,

$$Prob(o \mid s) = \prod_i Prob(o_i \mid s_i)$$

The channel model is trained by automatically aligning the hand transcriptions with the output of Sphinx-II on the utterances in the (SPEECHPP) training set and by tabulating the confusions that occurred. We use a Viterbi beam-search to find the $s$ that maximizes the expression. This technique is widely known so is not described here (see Forney (1973) and Lowerre (1986)).

Having a relatively small number of TRAINS-95 dialogues for training, we wanted to investigate how well the data could be employed in models for both the SR and the SPEECHPP. We ran several experiments to weigh our options. For a baseline, we built a class-based back-off language model for Sphinx-II using only transcriptions of ATIS spoken utterances. Using this model, the performance of Sphinx-II alone on TRAINS-95 data was 58.7%. Note that this figure is lower than our previously mentioned average of 70%, since we were unable to exactly replicate the ATIS model from CMU.

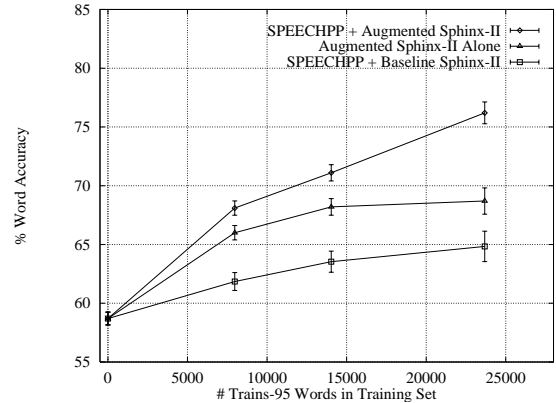

Figure 6: Post-processing Evaluation

First, we used varying amounts of training data exclusively for building models for the SPEECHPP; this scenario would be most relevant if the speech recognizer were a black-box and we did not know how to train its model(s). Second, we used varying amounts of the training data exclusively for augmenting the ATIS data to build language models for Sphinx-II. Third, we combined the methods, using the training data both to extend the language models for Sphinx-II and to then train SPEECHPP on the newly trained SR.

The results of the first experiment are shown by the bottom curve of Figure 6, which indicates the performance of the SPEECHPP with the baseline Sphinx-II. The first point comes from using approx. 25% of the available training data in the SPEECHPP models. The second and third points come from using approx. 50% and 75%, respectively, of the available training data. The curve clearly indicates that the SPEECHPP does a reasonable job of boosting our word recognition rates over baseline Sphinx-II and performance improves with additional training data. We did not train with all of our available data, since the remainder was used for testing to determine the results via repeated leave-one-out cross-validation. The error bars in the figure indicate 95% confidence intervals.

Similarly, the results of the second experiment are shown by the middle curve. The points reflect the performance of Sphinx-II (without SPEECHPP) when using 25%, 50%, and 75% of the available training data in its LM. These results indicate that equivalent amounts of training data can be used with greater impact in the language model of the SR than in the post-processor.

Finally, the outcome of the third experiment is reflected in the uppermost curve. Each point indicates the performance of the SPEECHPP using a set of models trained on the behavior of Sphinx-II for the corresponding point from the second experiment. The results from this experiment indicate that even if the language model of the SR can be modified, then the post-processor trained on the same new data can still significantly improve word recognition accuracy on a separate test set. Hence, whether the SR's models are tunable or not, the post-processor is in neither case redundant.

Since these experiments were performed, we have enhanced the channel model by relaxing the constraint that replacement errors be aligned on a word-by-word basis. We employ a fertility model (Brown et al, 1990) that indicates how likely each word is to map to multiple words or to a partial word in the SR output. This extension allows us to better handle the second example above, replacing TO TRY with DETROIT. For more details, see Ringger and Allen (1996).

## 5. Robust Parsing

Given that speech recognition errors are inevitable, robust parsing techniques are essential. We use a pure bottom-up parser (using the system described in (Allen, 1995)) in order to identify the possible constituents at any point in the utterance based on syntactic and semantic restrictions. Every constituent in each grammar rule specifies both a syntactic category and a semantic category, plus other features to encode co-occurance restrictions as found in many grammars. The semantic features encode selectional restrictions, most of which are domain-independent. For example, there is no general rule for PP attachment in the grammar. Rather there are rules for temporal adverbial modification (e.g., *at eight o'clock)*, locational modification (e.g., *in Chicago*), and so on.

The end result of parsing is a sequence of speech acts rather than a syntactic analysis. Viewing the output as a sequence of speech acts has significant impact on the form and style of the grammar. It forces an emphasis on encoding semantic and pragmatic features in the grammar. There are, for instance, numerous rules that encode specific conventional speech acts (e.g., *That's good* is a CONFIRM, *Okay* is a CONFIRM/ACKNOWLEDGE, *Let's go to Chicago* is a SUGGEST, and so on). Simply classifying such utterances as sentences would miss the point. Thus the parser computes a set of plausible speech act interpretation based on the surface form, similar to the model described in Hinkelman & Allen (1989).

We use a hierarchy of speech acts that encode different levels of vagueness, including a TELL act that indicates content without an identifiable illocutionary force. This allows us to always have an illocutionary force that can be refined as more of the utterance is processed. The final interpretation of an utterance is the sequence of speech acts that provides the "minimal covering" of the input - i.e., the shortest sequence that accounts for the input. Even if an utterance was completely uninterpretable, the parser would still produce output - a TELL act with no content.

For example, consider an utterance from the sample dialogue that was garbled: OKAY NOW I TAKE THE LAST TRAIN IN GO FROM ALBANY TO IS. The best sequence of speech acts to cover this input consists of three acts:
1. a CONFIRM/ACKNOWLEDGE (OKAY)
2. a TELL, with content to take the last train (NOW I TAKE THE LAST TRAIN)
3. a REQUEST to go from Albany (GO FROM ALBANY)

Note that the *to is* at the end of the utterance is simply ignored as it is uninterpretable. While not present in the output, the presence of unaccounted words will lower the parser's confidence score that it assigns to the interpretation.

The actual utterance was *Okay now let's take the last train and go from Albany to Milwaukee*. Note that while the parser is not able to reconstruct the complete intentions of the user, it has extracted enough to continue the dialogue in a reasonable fashion by invoking a clarification subdialogue. Specifically, it has correctly recognized the confirmation of the previous exchange (act 1), and recognized a request to move a train from Albany (act 3). Act 2 is an incorrect analysis, and results in the system generating a clarification question that the user ends up ignoring. Thus, as far as furthering the dialogue, the system has done reasonably well.

## 6. Robust Speech Act Processing

The dialogue manager is responsible for interpreting the speech acts in context, formulating responses, and maintaining the system's idea of the state of the discourse. It maintains a discourse state that consists of a goal stack with similarities to the plan stack of Litman & Allen (1987) and the attentional state of Grosz & Sidner (1986). Each element of the stack captures
1. the domain or discourse goal motivating the segment
2. the object focus and history list for the segment
3. information on the status of problem solving activity (e.g., has the goal been achieved yet or not).

A fundamental principle in the design of TRAINS-95 was a decision that, when faced with ambiguity it is better to choose a specific interpretation and run the risk of making a mistake as opposed to generating a clarification subdialogue. Of course, the success of this strategy depends on the system's ability to recognize and interpret subsequent corrections if they arise. Significant effort was made in the system to detect and handle a wide range of corrections, both in the grammar, the discourse processing and the domain reasoning. In later systems, we plan to specifically evaluate the effectiveness of this strategy.

The discourse processing is divided into reference resolution, verbal reasoning, problem solving and domain reasoning.

Reference resolution, other than having the obvious job of identifying the referents of noun phrases, also may reinterpret the parser's assignment of illocutionary force if it has additional information to draw upon. One way we attain robustness is by having overlapping realms of responsibility: one module may be able to do a better job resolving a problem because it has an alternative view of it. On the other hand, it's important to recognize another module's expertise as well. It could be disastrous to combine two speech acts that arise from *I really <garbled> think that's good.* for instance, since the garbled part may include *don't*. Since speech recognition may substitute important words one for the other, it's important to keep in mind that speech acts that have no firm illocutionary force due to grammatical problems may have little to do with what the speaker actually said.

The verbal reasoner is organized as a set of prioritized rules that match patterns in the input speech acts and the discourse state. These rules allow robust processing in the face of partial or ill-formed input as they match at varying levels of specificity, including rules that interpret fragments that have no identified illocutionary force. For instance, one rule would allow a fragment such as *to Avon* to be interpreted as a suggestion to extend a route, or an identification of a new goal. The prioritized rules are used in turn until an acceptable result is obtained.

The problem solver handles all speech acts that appear to be requests to constrain, extend or change the current plan. It is also based on a set of prioritized rules, this time dealing with plan corrections and extensions. These rules match against the speech act, the problem solving state, and the current state of the domain. If fragmentary information is supplied, the problem solver attempts to incorporate the fragment into what it knows about the current state of the plan.

As example of the discourse processing, consider how the system handles the user's first utterance in the dialogue, OKAY LET'S SEND CONTAIN FROM DETROIT TO WASHINGTON. From the parser we get three acts:
1. a CONFIRM/ACKNOWLEDGE (OKAY)
2. a TELL involving mostly uninterpretable words (LET'S SEND CONTAIN)
3. a TELL act that mentions a route (FROM DETROIT TO WASHINGTON)

The discourse manager sets up its initial conversation state and passes the act to reference for identification of particular objects, and then hands the acts to the verbal reasoner. Because there is nothing on the discourse stack, the initial confirm has no effect. (Had there been something on the stack, e.g. a question of a plan, the initial confirm might have been taken as an answer to the question, or a confirm of the plan, respectively). The following empty TELL act is uninterpretable and hence ignored. While it is possible to claim the "send" could be used to indicate the illocutionary force of the following fragment, and that a "container" might even be involved, the fact that the parser separated out the speech act indicates there may have been other fragments lost. The last speech act could be a suggestion of a new goal to move from Detroit to Washington. After checking that there is an engine at Detroit, this interpretation is accepted. The planner is unable to generate a path between these points (since it is greater than four hops). It returns two items:
1. an identification of the speech act as a suggestion of a goal to take a train from Detroit to Washington
2. a signal that it couldn't find a path to satisfy the goal

The discourse context is updated and the verbal reasoner generates a response to clarify the route desired, which is realized in the system's response *What route would you like to get from Detroit to Washington?*

As another example of robust processing, consider an interaction later in the dialogue in which the user's response *no* is misheard as *now*: *Now let's take the train from Detroit to Washington do S_X Albany* (instead of *No let's take the train from Detroit to Washington via Cincinnati*). Since no explicit rejection is identified due to the recognition error, this utterance looks like a confirm and continuation of the plan. Thus the problem solver is called to extend the path with the currently focused engine (engine1) from Detroit to Washington.

The problem solver realizes that engine1 isn't currently in Detroit, so this can't be a route extension. In addition, there is no other engine at Detroit, so this is not plausible as a focus shift to a different engine. Since engine1 originated in Detroit, it then decides to reinterpret the utterance as a correction. Since the utterance adds no new constraints, but there are the cities that were just mentioned as having delays, it presumes the user is attempting to avoid them, and invokes the domain reasoner to plan a new route avoiding the congested cities. The new path is returned and presented to the user.

While the response does not address the user's intention to go through Cincinnati due to the speech recognition errors, it is a reasonable response to the problem the user is trying to solve. In fact, the user decides to accept the proposed route and forget about going through Cincinnati. In other cases, the user might persevere and continue with another correction such as *No, through Cincinnati*. Robustness arises in the example because the system uses its knowledge of the domain to produce a reasonable response. Note these examples both illustrate the "strong commitment" model. We believe it is easier to correct a poor plan, than having to keep trying to explain a perfect one, particularly in the face of recognition problems. For further detail on the problem solver, see Ferguson et al (1996).

## 7. Evaluating the System

While examples can be illuminating, they don't address the issue of how well the system works overall. To explore how well the system robustly handles spoken dialogue, we designed an experiment to contrast speech input with keyboard input. The experiment uses the different input media to manipulate the word error rate and the degree of spontaneity. Task performance was evaluated in terms of two metrics: the amount of time taken to arrive at a solution and the quality of the solution. Solution quality for our domain is determined by the amount of time needed to travel the routes.

Sixteen subjects for the experiment were recruited from undergraduate computer science courses. None of the subjects had ever used the system before. The procedure was as follows:

- The subject viewed an online tutorial lasting 2.4 minutes.
- The subject was then allowed a few minutes to practice both speech and keyboard input.
- All subjects were given identical sets of 5 tasks to perform, in the same order. Half of the subjects were asked to use speech first, keyboard second, speech third and keyboard fourth. The other half used keyboard first and then alternated. All subjects were given a choice of whether to use speech or keyboard input to accomplish the final task.
- After performing the final task, the subject completed a questionnaire.

An analysis of the experiment results shows that the plans generated when speech input was used are of similar quality to those generated when keyboard input was used. However, the time needed to develop plans was significantly lower when speech input was used.

Overall, problems were solved using speech in 68% of the time needed to solve them using the keyboard. Figure 7 shows the task completion time results, and Figure 8 gives the solution quality results, each broken out by task.

Of the 16 subjects, 12 selected speech as the input medium for the final task and 4 selected keyboard input. Three of the four selecting keyboard input had actually experienced better or similar performance using keyboard input during the first four tasks. The fourth subject indicated on his questionnaire that he believed he could solve the problem more quickly using the keyboard; however, that subject had solved the two tasks using speech input 19% faster than the two tasks he solved using keyboard input.

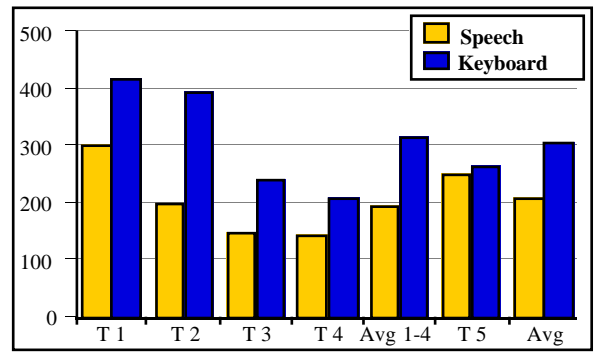

Figure 7: Time to Completion by Task

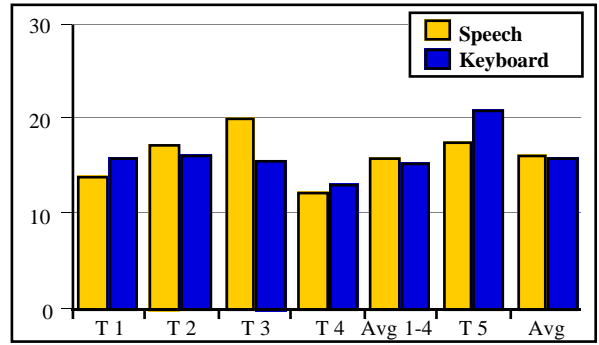

Figure 8 : Length of Solution by Task

Of the 80 tasks attempted, there were 7 in which the stated goals were not met. In each unsuccessful attempt, the subject was using speech input. There was no particular task that was troublesome and no particular subject that had difficulty. Seven different subjects had a task where the goals were not met, and each of the five tasks was left unaccomplished at least once.

A review of the transcripts for the unsuccessful attempts revealed that in three cases, the subject misinterpreted the system's actions, and ended the dialogue believing the goals were met. Each of the other four unsuccessful attempts resulted from a common sequence of events: after the system proposed an inefficient route, word recognition errors caused the system to misinterpret rejection of the proposed route as acceptance. The subsequent subdialogues intended to improve the route were interpreted to be extensions to the route, causing the route to "overshoot" the intended destination.

This suggests that, while our robustness techniques were effective on average, the errors do create a higher variance in the effectiveness of the interaction. These problems reveal a need for better handling of corrections, especially as resumptions of previous topics. More details on the evaluation can be found in (Sikorski & Allen, forthcoming).

## 8. Discussion

There are few systems that attempt to handle unconstrained natural dialogue. In most current speech systems, the interaction is driven by a template filling mechanism (e.g., the ATIS systems (ARPA, 1995), BeRP (Jurafsky et al, 1994), Pegasus (Seneff *et al*, 1995)). Some of these systems support system-initiated questions to elicit missing information in the template, but that is the extent of the mixed initiative interaction. Specifically, there is no need for goal management because the goal is fixed throughout the dialogue. In addition, there is little support for clarification and correction subdialogues. The Duke system (Smith and Hipp, 1994) uses a more general model based on a reasoning system, but allows only a limited vocabulary and grammar and requires extensive training to use.

Our approach here is clearly bottom-up. We have attempted to build a fully functional system in the simplest domain possible and focused on the problems that most significantly degraded overall performance. This leaves us open to the criticism that we are not using the most sophisticated models available. For instance, consider our generation strategy. Template-based generation is clearly inadequate for many generation tasks. In fact, when starting the project we thought generation would be a major problem. However, the problems we expected have not arisen. While we could clearly improve the output of the system even in this small domain, the current generator does not appear to drag the system's performance down. We approached other problems similarly. We tried the simplest approaches first and then only generalized those algorithms whose inadequacies clearly degrade the performance of the system.

Likewise, we view the evaluation as only a very preliminary first step. While our evaluation appears similar to HCI experiments on whether speech or keyboard is a more effective interface in general (cf. Oviatt and Cohen, 1991), this comparison was not our goal. Rather, we used the modality switch as a way of manipulating the error rate and the degree of spontaneity. While keyboard performance is not perfect because of typos (we had a 5% word error rate on keyboard), it is considerably less error prone than speech. All we conclude from this experiment is that our robust processing techniques are sufficiently good that speech is a viable interface in such tasks even with high word error rates. In fact, it appears to be more efficient in this application than keyboard. In contrast to the results of Rudnicky (1993), who found users preferred speech even when less efficient, our subjects generally preferred the most efficient modality for them (which in a majority of cases was speech).

Despite the limitations of the current evaluation, we are encouraged by this first step. It seems obvious to us that progress in dialogue systems is intimately tied to finding suitable evaluation measures. And task-based evaluation seems one of the most promising candidates. It measures the impact of proposed techniques directly rather than indirectly with an abstract accuracy figure.

Another area where we are open to criticism is that we used algorithms specific to the domain in order to produce effective intention recognition, disambiguation, and domain planning. Thus, the success of the system may be a result of the domain and say little about the plan-based approach to dialogue. To be honest, with the current system, it is hard to defend ourselves against this. This is is a first step in what we see as a long ongoing process. To look at it another way: if we couldn't build a successful system by employing whatever means available, then there is little hope for finding more effective general solutions.

We are addressing this problem in our current research: we are developing a domain-independent plan reasoning "shell" that manages the plan recognition, evaluation and construction around which the dialogue system is structured. This shell provides the abstract model of problem solving upon which the dialogue manager is built. It is then instantiated by domain specific reasoning algorithms to perform the actual searches, constraint checking and intention recognition for a specific application. The structure of the model remains constant across domains, but the actual details of constructing plans remain domain specific.

Our next iteration of this process, TRAINS-96, involves adding complexity to the dialogues by increasing the complexity of the task. Specifically, we are adding distances and travel times between cities, several new modes of transportation (trucks and planes) with associated costs, and simple cargoes to be transported and possibly transferred between different vehicles. The expanded domain will require a much more sophisticated ability to answer questions, to display complex information concisely, and will stress our abilities to track plans and identify focus shifts.

While there are clearly many places in which our current system requires further work, it does set a new standard for spoken dialogue systems. More importantly, it allows us to address new research issues in a much more systematic way, supported by empirical evaluation.

## Acknowledgements

This work was supported in part by ONR/ARPA grants N0004-92-J-1512 and N00014-95-1-1088, and NSF grant IRI-9503312. Many thanks to Alex Rudnicky, Ronald Rosenfeld and Sunil Issar at CMU for providing the Sphinx-II system and related tools. This work would not have been possible without the efforts of George Ferguson on the TRAINS system infrastructure and model of problem solving.

## References

J. F. Allen. 1995. *Natural Language Understanding*, *2nd Edition*, Benjamin-Cummings, Redwood City, CA.

J. F. Allen, G. Ferguson, B. Miller, and E. Ringger. 1995. Spoken dialogue and interactive planning. In *Proc*. ARPA SLST Workshop, Morgan Kaufmann

J. F. Allen and C. R. Perrault. 1980. Analyzing intention in utterances, *Artificial Intelligence* 15(3):143-178

ARPA, 1995. Proceedings of the Spoken Language Systems Technology Workshop, Jan. 1995. Distributed by Morgan Kaufmann.

P. F. Brown, J. Cocke, S. A. Della Pietra, V. J. Della Pietra, F. Jelinek, J. D. Lafferty, R. L. Mercer and P. S. Roossin. 1990. A Statistical Approach to Machine Translation. C*omputational Linguistics* 16(2):79--85.

S. Carberry. 1990. *Plan Recognition in Natural Language Dialogue*, MIT Press, Cambridge, MA.

P. R. Cohen and C. R. Perrault. 1979. Elements of a plan-based theory of speech acts, *Cognitive Science* 3

G. M. Ferguson, J. F. Allen and B. W. Miller, 1996. TRAINS-95: Towards a Mixed-Initiative Planning Assistant, to appear in *Proc. Third Conference on Artificial Intelligent Planning Systems (AIPS-96)*.

G. E. Forney, Jr. 1973. The Viterbi Algorithm. Proc. of IEEE 61:266--278.

B. Grosz and C. Sidner. 1986. Attention, intention and the structure of discourse. *Computational Linguistics* 12(3).

E. Hinkelman and J. F. Allen. 1989.Two Constraints on Speech Act Ambiguity, Proc. ACL.

X. D. Huang, F. Alleva, H. W. Hon, M. Y. Hwang, K. F. Lee, and R. Rosenfeld. 1993. The Sphinx-II Speech Recognition System. *Computer, Speech and Language*

D. Jurafsky, C. Wooters, G. Tajchman, J. Segal, A. Stolcke, E. Fosler and N. Morgan. 1994. The Berkeley Restaurant Project, *Proc. ICSLP-94*.

S. M. Katz. 1987. Estimation of Probabilities from Sparse Data for the Language Model Component of a Speech Recognizer. In *IEEE Transactions on Acoustics, Speech, and Signal Processing*. IEEE. pp. 400-401.

D. Litman and J. F. Allen. 1987. A plan recognition model for subdialogues in conversation. *Cognitive Science* 11(2): 163-200

B, Lowerre and R. Reddy. 1986. The Harpy Speech Understanding System. Reprinted in Waibel and Lee, 1990: 576-586.

S. L. Oviatt and P.R. Cohen. 1991. The contributing influence of speech and interaction on human discourse patterns. In J.W. Sullivan and S.W. Tyler (eds), *Intelligent User Interfaces*. Addison-Wesley, NY, NY.

E. K. Ringger and J. F. Allen. 1996. A Fertility Channel Model for Post-Correction of Continuous Speech Recognition. To appear in *Proc. 1996 ICSLP*, IEEE, October, 1996.

A. Rudnicky. 1993. Mode Preference in a Simple Data-Retrieval Task, *Proc. of ARPA Workshop on Human Language Technology*, Dist. by Morgan Kaufmann.

S. Seneff, V. Zue, J. Polifroni, C. Pao, L. Hetherington, D. Goddeau, and J. Glass. 1995. The Preliminary Development of a Displayless PEGASUS System. *Proc.* SLST Workshop, Jan. 1995. Morgan Kaufmann

R. Smith and R. D. Hipp. 1994. *Spoken Natural Language Dialog Systems: A Practical Approach*, Oxford University Press.

A. Waibel and K. F. Lee, editors. 1990. *Readings in Speech Recognition*. Morgan Kaufmann, CA.

# Appendix A: Transcript of post-processor repairs in the dialogue.

HYP: OKAY LET'S SEE CONTAIN FROM DETROIT TO WASHINGTON
HYP': OKAY LET'S SEND CONTAIN FROM DETROIT TO WASHINGTON
REF: OKAY LET'S TAKE THE TRAIN FROM DETROIT TO WASHINGTON

HYP: LET'S GO VIA TOLEDO AND PITTSBURGH
HYP': LET'S GO VIA TOLEDO AND PITTSBURGH
REF: LET'S GO VIA TOLEDO AND PITTSBURGH

HYP: NOW LET'S TAKE THE TRAIN FROM DETROIT TO WASHINGTON D S_X ANY
HYP': NOW LET'S TAKE THE TRAIN FROM DETROIT TO WASHINGTON DO S_X ALBANY)
REF: NO LET'S TAKE THE TRAIN FROM DETROIT TO WASHINGTON VIA CINCINNATI

HYP: OKAY IT THAT'S OKAY NOW
HYP': OKAY IT THAT'S OKAY NOW
REF: OKAY THAT- THAT'S OKAY NOW

HYP: UP OKAY NOW LET'S TAKE THE TRAIN FROM MONTREAL TO LEXINGTON
HYP': UP OKAY NOW LET'S TAKE THE TRAIN FROM MONTREAL TO LEXINGTON
REF: OKAY NOW LET'S TAKE THE TRAIN FROM MONTREAL TO LEXINGTON

HYP: LET'S GO P_M TO TRY
HYP': LET'S GO P_M TO DETROIT
REF: LET'S GO VIA DETROIT

HYP: YES NOW LET'S GO TO LEXINGTON
HYP': YES NOW LET'S GO TO LEXINGTON
REF: YES NOW LET'S GO TO LEXINGTON

HYP: OKAY NOW ADD STATE THE LAST TRAIN AND GO FROM ALBANY TO O_O'S
HYP': OKAY NOW I TAKE THE LAST TRAIN IN GO FROM ALBANY TO IS
REF: OKAY NOW LET'S TAKE THE LAST TRAIN AND GO FROM ALBANY TO MILWAUKEE

HYP: ME A JET ADD ALBANY NEEDS TO GO TO MILWAUKEE
HYP': ME THE JET AT ALBANY INSTEAD TO GO TO MILWAUKEE
REF: THE ENGINE AT ALBANY NEEDS TO GO TO MILWAUKEE

HYP: I'D GO VIA BUFFALO
HYP': UH GO VIA BUFFALO
REF: UH GO VIA BUFFALO

HYP: GO B_X SYRACUSE AT BUFFALO
HYP': GO VIA SYRACUSE VIA BUFFALO
REF: GO VIA SYRACUSE AND BUFFALO

HYP: THAT'S COULD I CAN
HYP': THAT'S GREAT UH CAN
REF: THAT'S GOOD I'M DONE

HYP: I_NEED DONE
HYP': I'M DONE
REF: I'M DONE

# Appendix B: A Keyboard Dialogue

U:     Let's take the train in Charlotte to Lexington
S:     Yeah
U:     Can we go through Atlanta instead?
S:     Yes
U:     much better
S:     Oh?
U:     Let's move from Toronto to Washington
S:     What route would you like to get from Toronto to Washington?
U:     through Montreal.
S:     Yes. Is this ok?
U     good, now can we go down through Albany
S:     Huh?
U:     from Montreal, let's take the train south through Burlington and Albany.
S:     OK. Is this ok?
U:     Yes, now let's move the train in Boston to Washington.
S: OK. City New York is congested due to unusually heavy traffic. An additional 5 hours will be needed to travel through there.
U:     let's go around New York instead.
S: OK
U: How about going through Albany.
S: Yep
U: I'm sorry, can we move the train in Boston to Philadelphiainstead (sic).
S:     It would take a far more powerful and expensive computer than me to figure the answer to that one out.
U:     I'm sorry, can we move the train from Boston to Philadelphia.
S:     OK. But, Your routes cross at City Baltimore. Trains will take an additional 5 hours to move through the crossed cities.
U:     Let's redo the route for Boston.
S:     What was that?
U:     Let's clear Boston's current route.
S:     Yeah
U:     Let's clear Boston's current route.
S:     Yes
U:     Let's move from Boston to Philadelphia
S:     OK
U:     I think we are done.